%% file: mainpaper.tex
\documentclass[runningheads]{llncs}
\usepackage[T1]{fontenc}
\usepackage{hyperref}
\usepackage{graphicx,verbatim}
\usepackage{color}

\usepackage{amssymb}
\usepackage{amsmath}
 \usepackage{multirow}
 \usepackage{rotating}
\usepackage[dvipsnames]{xcolor}
\begin{document}

\newcommand{\bibek}[1]{\textcolor{magenta}{#1}}
\title{ProMoE-FL: Prototype-conditioned Mixture of Experts for Multimodal Federated Learning with Missing Modalities}
% \titlerunning{ProMoE-FL: Prototype cond. MoE for MMFL with Missing Modalities}
\titlerunning{ProMoE-FL}
% If the paper title is too long for the running head, you can set
% an abbreviated paper title here
%

\author{Aavash Chhetri\inst{1} \and
Bibek Niroula\inst{1} \and
Eduard Vazquez\inst{2} \and
Yash Raj Shrestha\inst{3} \and
Prashnna Gyawali\inst{4} \and
Loris Bazzani\inst{5} \and
Binod Bhattarai\inst{1,6,7} 
}
% \index{Chhetri, Aavash}
% \index{Niroula, Bibek}
% \index{Vazquez, Eduard}
% \index{Shrestha, Yash Raj}
% \index{Gyawali, Prashnna}
% \index{Bazzani, Loris}
% \index{Bhattarai, Binod}
\authorrunning{A. Chhetri et al.}
% First names are abbreviated in the running head.
% If there are more than two authors, 'et al.' is used.
%
\institute{NepAl Applied Mathematics and Informatics Institute for research, Nepal \and
Fogsphere (Redev.AI Ltd), UK \and
University of Lausanne, Switzerland \and
West Virginia University, USA \and
University of Verona, Italy \and
University College London, UK \and 
University of Aberdeen, UK \\
\email{binod.bhattarai@abdn.ac.uk}}
% \begin{comment}
% \end{comment}

% \author{Anonymized Authors}  %% Added for anonymized MICCAI submission
% \authorrunning{Anonymized Author et al.}
% \institute{Anonymized Affiliations \\
%     \email{email@anonymized.com}}
  
\maketitle              % typeset the header of the contribution
\begin{abstract}
In this paper, we address the problem of multimodal federated learning with missing modality.
Existing methods utilize an additional public dataset or perform naive feature synthesis that is based solely on the available modality.
To address these limitations, we propose ProMoE-FL, a Prototype-conditioned Mixture-of-Experts framework for robust missing-modality feature synthesis in multimodal federated learning. ProMoE-FL builds a global client-aware prototype bank that captures clinically meaningful modality priors across institutions. 
Our Mixture of Experts is conditioned on these prototypes and modality indices to enable direction-aware expert routing for dynamically synthesizing missing features.
We perform extensive quantitative and qualitative evaluations on four public chest X-ray datasets (MIMIC-CXR, NIH Open-I, PadChest, and CheXpert) and demonstrate that ProMoE-FL consistently outperforms state-of-the-art methods in both homogeneous as well as the more challenging heterogeneous settings.
Code available at: \href{https://github.com/bhattarailab/ProMoE-FL}{https://github.com/bhattarailab/ProMoE-FL}.
% Authors must provide keywords and are not allowed to remove this Keyword section.

\end{abstract}

\section{Introduction}
%%% Introduction will also explain related works under the same section
%why multimodal learning in healthcare
Clinicians inherently integrate information from a variety of sources and modalities such as pathology reports, laboratory data, omics data, X-Ray, Magnetic Resonance Imaging (MRI), and Computed Tomographhy (CT) to form holistic and context-aware clinical judgements. This has motivated the recent development of models that make use of highly multimodal data to computationally mirror such integrative clinical reasoning~\cite{Acosta2022Multimodal,jmirmultimodal,Shrestha2023Medical}.
%why federated learning
In healthcare, multimodal data is inherently fragmented across medical centers due to privacy regulations and data governance constraints, posing significant challenges to centralized training paradigms. 
Federated Learning (FL)~\cite{McMahan2017Communication} addresses these challenges and provides a privacy-preserving approach to collectively training shared healthcare models without any exchange or distribution of patient data residing in medical centers. Recent efforts have therefore explored multimodal learning under the federated setting~\cite{Sachin2023AMC,9891834qayyum,chenReportgen}.
%missing modalities in multimodal learning
However, most of the existing literature on multimodal federated learning in healthcare assumes that all participating institutions have access to complete sets of modalities \cite{Thrasher2023Multimodal,chhetri2026medmmflmultimodalfederatedlearning}. In real-world settings, multimodal clinical data are frequently incomplete across centres owing to disparities in equipment availability, acquisition protocols, and data management resources.
This challenge calls for novel methodologies for dealing with missing modalities.
%approaches to solve missing modalities
In centralized multimodal learning, the challenge of missing modalities is addressed through prompt-based \cite{Seibold_2022,You_2023}, generative \cite{chen2023generativetextguided3dvisionlanguage,lee2024visionlanguagegenerativemodelviewspecific,dai2025unbiased}, dropout-based \cite{lau2019unifiedrepresentation} methods, or utilize specialized architectures \cite{MMIN,ShaSpec,SMIL} to model inter-modal dependencies. 
Such centralized approaches are impractical in FL scenarios, as their 
architectural designs do not consider the strict privacy, communication efficiency, and heterogeneity requirements inherent to practical FL deployments.

%approaches to solve missing modalities in Federated Settings
More recently, a growing body of FL work has begun to address the challenge of missing modalities, predominantly in non-medical domains~\cite{yu2023multimodalfederatedlearningcontrastive,LE2025103219,bao2024multimodalfederatedlearningmissing,fedcolaSun}, 
while a limited number of studies were proposed for the medical domain~\cite{Pou_CARMFL_MICCAI2024,poudel2025multimodalfederatedlearningmissing,saha2024examiningmodalityincongruitymultimodal,Saha_Mishra_Wagner_Kamnitsas_Noble_2025}.
%medical domain works
CAR-MFL \cite{Pou_CARMFL_MICCAI2024} handles missing modalities by training with additional public multimodal data to compensate for incomplete client modalities. However, this strategy assumes the availability of representative public data, which is often unrealistic in healthcare due to privacy constraints and domain gaps between public and private datasets.
FeatImp \cite{poudel2025multimodalfederatedlearningmissing} performs feature-level imputation by training separate networks to learn modality-specific mappings to reconstruct missing features.
This approach assumes stable cross-modal relationships across clients and synthesizes features solely from the observed modality, without explicitly modeling complementary information inherent to the missing modality.
PmcmFL~\cite{bao2024multimodalfederatedlearningmissing} substitutes class-level prototypes as features of missing modalities. Although prototype-based and closer to our formulation, it assigns identical representations to all samples within a class which limits instance-level diversity and fails to capture intra-class variability, which is critical in medical data.

%our contribution
%shorter version
To address these limitations, we propose ProMoE-FL, a prototype-conditioned Mixture-of-Experts framework for missing modality synthesis in federated learning. The method generates modality-specific features by conditioning on available modalities and a global prototype bank of missing modalities. This enables privacy-preserving and structured feature reconstruction without external data. A shared, routing-based Mixture-of-Experts (MoE) architecture allows experts to be reused across modalities, preventing combinatorial parameter growth.
Furthermore, by accumulating client-specific prototypes that summarize local modality distributions, the global prototype bank serves as a client-aware prior in the latent space. This guides feature synthesis to account for distributional variability across clients, thereby improving robustness under the non-independently and identically distributed (i.e., non-IID) data characteristics prevalent in real-world federated settings.
Our key contributions are the following:
\begin{itemize}
    \item We propose ProMoE-FL, a prototype-conditioned feature generation framework for missing modality feature synthesis in federated multimodal learning.
    \item We introduce client-aware prototype attention to address cross-client heterogeneity and employ a MoE architecture for direction-aware synthesis within a shared parameterization.
    \item We conduct extensive experiments under heterogeneous settings that mirror real-world clinical data silos, demonstrating improved robustness. We further support our findings with qualitative analyses.
\end{itemize}

\section{Method}

%multimodal federated learning formulation
We formulate our method for a typical multimodal federated setting with $K$ clients, each with its private dataset $D_k$ containing $n_k$ samples. 
The $n$-th data sample in $D_k$ is represented by the tuple $(\{X_m^{(n)}\}_{m=1}^{M_k},Y^{(n)})$, where $Y^{(n)}$ denotes the label related to $C$ classes and $M_k$ is the number of modalities available at the $k$-th client. 
All clients share common architecture with the global model, which is defined as $\{\boldsymbol{w} =  \{f_e^{m}\}_{m \in \mathcal{M}}, \; f_c\}$, where $\mathcal{M}$ denotes the complete set of modalities, $f_e^{m}$ is the encoder for modality $m$, and $f_c$ is the classifier.
We denote the latent feature representation extracted by the modality specific encoder as
$h_m^{(n)} = f_e^{m}(X_m^{(n)}), \quad m \in \mathcal{M}$.
%Global and Local objectives
The objective of federated training is to minimize the data-weighted empirical risk across all the clients, which is expressed as:
\begin{equation}
\boldsymbol{w}^* = \arg\min_{\boldsymbol{w}} \sum_{k=1}^{K} \frac{|D_k|}{|D|} \, \mathcal{L}_{task}^{(k)}(\boldsymbol{w}),
\label{eq:global_optimization}
\end{equation}
with the local loss at client $k$:
\begin{equation}
\mathcal{L}_{task}^{(k)}(\boldsymbol{w}) =
\frac{1}{|D_k|}
\sum_{n=1}^{|D_k|}
\mathcal{L}_{task}
\Big(
f_c\big(
\bigoplus_{m \in \mathcal{M}} f_e^{m}(X_m^{(n)})
\big),
Y^{(n)}
\Big),
\label{eq:local_optimization}
\end{equation}
where, $\oplus$ denotes a feature fusion operator, implemented as concatenation of modality-specific latent representations, and $D = \bigcup_{k=1}^{K} D_k$.
%Issue of missing modalities
In practical multi-center clinical deployments, multimodal systems frequently face missing modalities. A common approach to handle missing modality is zero-filling. 
For example, if the image modality is unavailable while text is present, the objective reduces to 
$\mathcal{L}_{task}\big(f_c(0 \oplus h_T^{(n)}), Y^{(n)}\big)$. 
Naively optimizing it distorts the joint feature space and biases the global model toward frequently observed modalities, thereby under-representing sparse ones.

%method figure
\begin{figure}[t]
    \centering
    \includegraphics[width=1\linewidth]{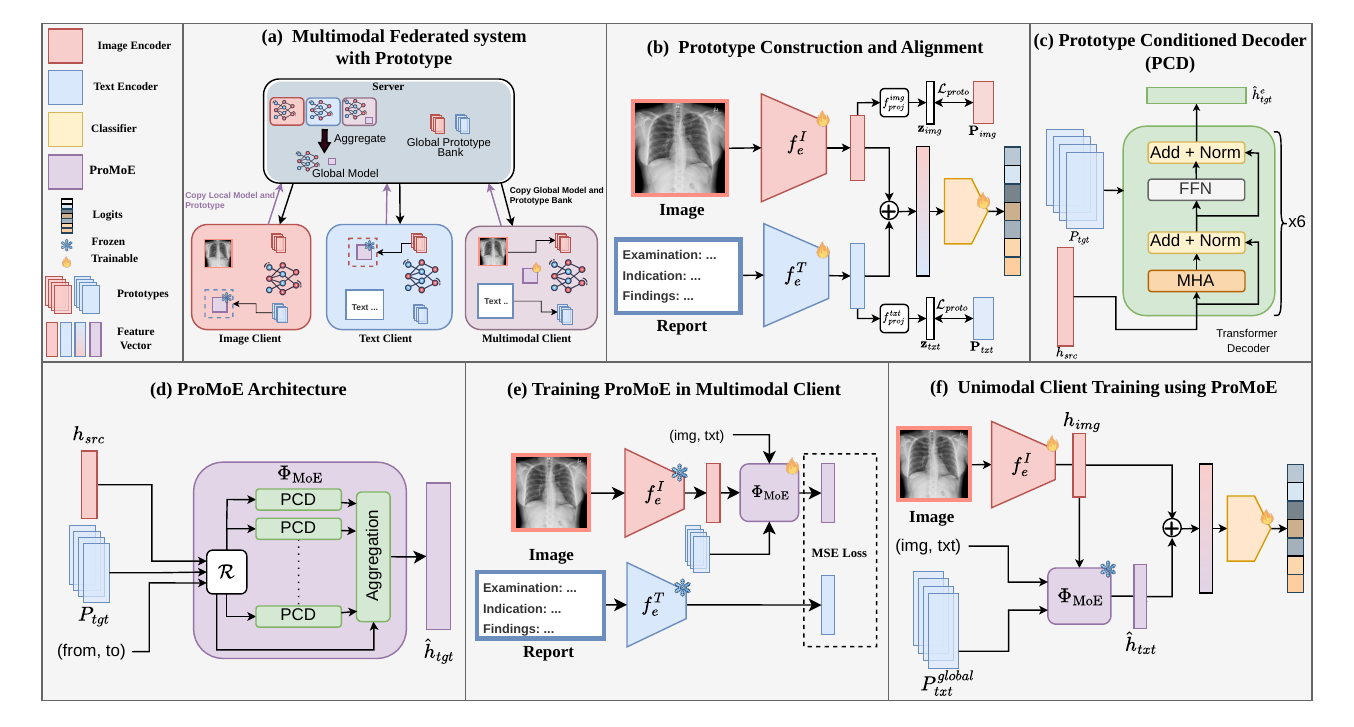}
    \caption{Overview of the proposed ProMoE-FL framework.}
% (a) Multimodal federated learning system with different types of client. 
% (b) Prototype construction and alignment via learnable modality-specific prototypes. 
% (c) Architecture of the Prototype-Conditioned Decoder (PCD). 
% (d) ProMoE architecture integrating multiple PCD experts. 
% (e) Training of ProMoE on multimodal clients using available image-text feature pairs. 
% (f) Unimodal client training, where missing modality features are synthesized using the ProMoE component.
    \label{fig:method_diagram}
\end{figure}

%high level overview of our approach
Fig.~\ref{fig:method_diagram} presents the overall framework of ProMoE-FL. For unimodal clients, a feature synthesis network reconstructs representations of missing modalities. 
While our approach operates in the feature space similarly to~\cite{poudel2025multimodalfederatedlearningmissing}, ProMoE-FL differs in both design and robustness.
Learning modality mappings solely from multimodal clients can degrade performance under client heterogeneity, where cross-modal relationships vary across sites and available modalities provide incomplete priors. To mitigate this, we introduce a client-aware global prototype bank encoding federation-wide modality priors, conditioning the synthesis process on target-modality prototypes.
%to improve robustness against distribution shifts. 
Furthermore, to ensure scalability without combinatorial parameter growth, the synthesis module is implemented as a Mixture-of-Experts architecture that dynamically routes modality-specific transformations through expert subnetworks.

%Prototype Component
\noindent \textbf{Prototype Construction and Alignment.} 
%motivation
In heterogeneous federated settings, modality embeddings often diverge across clients due to distribution shifts and inconsistent modality availability~\cite{li2021modelcontrastivefederatedlearning}. As a result, mappings learned only from multimodal clients may be insufficient for reliable cross-modal synthesis. Inspired by~\cite{LE2025103219,bao2024multimodalfederatedlearningmissing}, we introduce modality-specific, client-aware prototypes that act as global priors of the missing modality to complement the available modality and improve robustness to cross-client heterogeneity.
%working mechanism
For each modality $m$, every client $k$ maintains a set of $C$ learnable prototypes 
$\mathbf{P}_m^k = \{\mathbf{p}_1^{(m,k)}, \ldots, \mathbf{p}_C^{(m,k)}\}$ 
with $\mathbf{p}_c^{(m,k)} \in \mathbb{R}^d$. 
%shortened version
For each sample with at least one positive label, the modality-specific target centroid $\mathbf{c}_m^{(n)}$ is obtained by aggregating the corresponding class prototypes weighted by the multi-hot label vector and normalizing the result to unit norm.
For each modality $m$, we obtain a projected representation 
$\mathbf{z}_m^{(n)} = f_{\mathrm{proj}}^{m}(\mathbf{h}_m^{(n)}) \in \mathbb{R}^d$ 
using a modality-specific projection head implemented as a Fully Connected (FC) layer, and it is aligned with its corresponding centroid via
$\mathcal{L}_{proto}^{(m)} =
\frac{1}{|\mathcal{I}|}
\sum_{n \in \mathcal{I}}
\|\mathbf{z}_m^{(n)} - \mathbf{c}_m^{(n)}\|_2^2$,
where $\mathcal{I} = \{ n \mid \sum_{c=1}^{C} y_c^{(n)} > 0 \}$.
As illustrated in Fig.~\ref{fig:method_diagram}(b), we jointly optimize the client parameters $\mathbf{w}^k$ together with 
modality-specific projection heads and prototypes,
$\{ f_{\mathrm{proj}}^{m}, \mathbf{P}_k^{m} \}_{m \in \mathcal{M}_k}$, 
where $\mathcal{M}_k \subseteq \mathcal{M}$ denotes the set of modalities available on client $k$.
This is done via the following full training objective:
\begin{equation}
\label{eq:local_loss}
\mathcal{L} = 
\mathcal{L}_{\mathrm{task}} + 
\lambda \sum_{m \in \mathcal{M}_k} \mathcal{L}_{\mathrm{proto}}^{(m)},
\end{equation}
where $\mathcal{L}_{\mathrm{task}}$ is the task-specific loss (Eq.~\ref{eq:local_optimization}), 
$\mathcal{L}_{\mathrm{proto}}^{(m)}$ is the prototype alignment loss for modality $m$, 
and $\lambda$ controls the relative importance of the prototype loss.  
The prototypes are updated through gradient-based optimization each communication round, which encourages alignment with class representations while maintaining diversity, hence, improving robustness in heterogeneous settings. 

% Synthesis network component
\noindent \textbf{Missing Modality Feature Synthesis.}
%Prototype-conditioned Decoder (PCD)
For unimodal clients, we propose a \textit{Prototype-Conditioned Decoder} (PCD) $\phi$, illustrated in Fig.~\ref{fig:method_diagram}(c), to enable cross-modal feature synthesis. 
The PCD is implemented as a Transformer decoder where the source feature $h_{i}$ of available modality $i$ attends over the global prototype bank $P_{j}$ of the target modality $j$ received from the server to retrieve context necessary for synthesizing the target modality feature. 
Importantly, $P_{j}$ is constructed via accumulation of prototypes from clients rather than aggregation (e.g., averaging). 
Consequently, the decoder can dynamically attend to the most relevant target prototypes during synthesis, enabling adaptive complementary knowledge retrieval across heterogeneous domains.

%MOE
However, a single PCD may be insufficient to capture complex, non-linear relationships between heterogeneous modality pairs, especially in highly multimodal scenarios in the medical domain~\cite{chhetri2026medmmflmultimodalfederatedlearning}. 
Thus, we propose a Mixture of Experts $\Phi_{\mathrm{MoE}}$, illustrated in Fig.~\ref{fig:method_diagram}(d) where we define $E$ PCDs $\phi_e(h_{i}, P_{j})$ as independent experts $\{\phi_e\}_{e=1}^E$.
In order to redirect the data flow via the right experts, a Modality-Aware Router $\mathcal{R}$ computes a gating distribution conditioned on the instance context and the modality indices. Therefore, the network synthesizes the bottleneck feature of the missing modality $j$ as:
\begin{equation}
\hat{h}_j^{(n)}
=
\Phi_{\mathrm{MoE}}(h_i^{(n)}, P_j; i, j)
=
\frac{
\sum_{e=1}^{E}
\mathcal{R}(h_i^{(n)}, P_j, i, j)_e \;
\phi_e(h_i^{(n)}, P_j)
}{
\left\|
\sum_{e=1}^{E}
\mathcal{R}(h_i^{(n)}, P_j, i, j)_e \;
\phi_e(h_i^{(n)}, P_j)
\right\|_2
}.
\end{equation}
This formulation allows $\Phi_{\mathrm{MoE}}$ to adaptively select the optimal combination of prototype-conditioned experts for any missing modality scenario.
Although the proposed framework is formulated for an arbitrary set of modalities $\mathcal{M}$, for clarity of exposition we specialize the following derivations to the two-modality case, $\mathcal{M} = \{I, T\}$ used in our experiments, namely Image ($I$) and Text ($T$), without loss of generality.
For training unimodal clients, as shown in Fig.~\ref{fig:method_diagram}(f), the task-calibrated loss is hence, defined as:
\begin{equation}
\mathcal{L}_{task}
=
\mathcal{L}\!\left(
fc\!\left(
h_i^{(n)} \oplus \hat{h}_j^{(n)}
\right),
Y^{(n)}
\right),
\qquad
\hat{h}_j^{(n)} = \Phi_{\mathrm{MoE}}(h_i^{(n)}, P_j; i, j)
\end{equation}
where, $i,j \in \mathcal{M},\; i \neq j.$

\noindent \textbf{Federated Training in ProMoE-FL.}
At the start of each communication round, the server dispatches $\Phi_{\mathrm{MoE}}$ , 
% the complete global model 
$\boldsymbol{w}$, and prototype bank $\mathbf\{{P}_m^{global}\}_{m \in \mathcal{M}}$ to all clients.
Each client performs local training for $N$ epochs,
% on its private data, 
optimizing the training objective $\mathcal{L}$ as defined in Eq.~\ref{eq:local_loss}.
%MoE training
On multimodal clients, we construct a pool of paired features
$\mathcal{H}_k = \{(h_I^{(n)}, h_T^{(n)}) \mid (x_I^{(n)}, x_T^{(n)}) \in D_k\}$.
%alternative wording, timing is implied by then
As illustrated in~\ref{fig:method_diagram}(e), this pool is then used to train the feature synthesis network $\Phi_{MoE}^k$ using a symmetric reconstruction objective:
\begin{equation}
\mathcal{L}(\Phi_{MoE}^k)
=
\frac{1}{|\mathcal{H}_k|}
\sum_{(h_I^{(n)}, h_T^{(n)}) \in \mathcal{H}_k}
\sum_{\substack{i,j \in \mathcal{M}_k\\ i \neq j}}
\left\|
\Phi_{MoE}^k(h_{i}^{(n)}, P_{j}; i, j)
-
h_{j}^{(n)}
\right\|_2^2 .
\end{equation}
At the end of each communication round, each client transmits its updated model parameters $\boldsymbol{w}^k$, the modality-specific projection heads, 
and the local prototype sets $\mathbf{P}_m^k$ to the server. Multimodal clients additionally communicate $\Phi_{MoE}^k$. The resulting communication overhead is negligible, as the prototype bank, projection, and modality-aware routing mechanism together account for only approximately $0.3\%$ of the total model parameters. All the communicated parameters are aggregated via FedAvg~\cite{McMahan2017Communication} due to its simplicity and effectiveness as demonstrated by prior studies \cite{poudel2025multimodalfederatedlearningmissing,Pou_CARMFL_MICCAI2024,saha2024examiningmodalityincongruitymultimodal}. In contrast to the model parameters, the server accumulates all client prototypes into a global prototype bank: 
$\mathbf{P}_m^{global} =
\bigcup_{k=1}^{K} \mathbf{P}_m^k$ for each modality $m \in \mathcal{M}.$

\section{Experiments and Results}
\noindent \textbf{Datasets and Setups.}
%datasets
We perform a thorough experimentation on challenging datasets in both homogeneous and heterogeneous setups: MIMIC-CXR~\cite{johnson2019mimic,goldberger2000physionet}, NIH Open-I~\cite{demner2016preparing}, PadChest~\cite{BUSTOS2020101797Padchest}, and CheXpert \cite{irvin2019chexpert}.
We use the validation and test splits provided by MIMIC-CXR for the evaluation of the global model, following the protocol outlined in~\cite{Pou_CARMFL_MICCAI2024}.
We experimented with a variety of client configurations, denoted as \textbf{I:T:M}, where I, T, and M represent the numbers of image-only, text-only, and multimodal clients, respectively.
%homogenous setup
The homogenous setup comprises 10 clients, each with training data for 810 patients sampled from the same dataset MIMIC-CXR. 
%heterogenous setup
In the heterogeneous setup, 
multimodal clients include 2 clients from NIH Open-I, while any additional multimodal clients are sampled from PadChest. 
All image-only and text-only clients are drawn from CheXpert and PadChest, respectively.\\
\noindent \textbf{Implementation Details.}
We employ pretrained Resnet50 \cite{he2016deep} and BERT-base \cite{devlin2018bert} architecture as the image and text encoders, respectively. 
Their outputs are projected to 256-dimensional features and L2-normalized prior to concatenation. 
Optimization is performed using Adam \cite{kingma2014adam} with a learning rate of $1e^{-4}$.
$\lambda$ is tuned from $\{0.1, 0.5, 1, 5, 10\}$.
Each client trains locally for 3 epochs per communication round over 30 rounds in total.
Performance is measured using macro AUC (Average area under the Receiver Operating Characteristic Curve).
The results are averaged over three random seeds.
\\
\noindent \textbf{Baselines.}
We compare our method with two simple yet effective baselines: Zero-filling which replaces missing modalities with zero vectors and Uniform-filling which uniformly samples feature vectors (FedAvg \cite{McMahan2017Communication} + Zero/Unif.). 
We compare with state-of-the-art (SOTA) methods: FeatImp \cite{poudel2025multimodalfederatedlearningmissing} and PmcmFL \cite{bao2024multimodalfederatedlearningmissing}, representing feature-level imputation and prototype-based approaches, respectively.
One of the advantages of our ProMoE-FL is that it does not rely on public data like CAR-MFL \cite{Pou_CARMFL_MICCAI2024}.
However, ProMoE-FL can be easily adapted by incorporating  a virtual client that uses public data during training. 
In this way, we can fairly compare with public-data-based SOTA, CAR-MFL \cite{Pou_CARMFL_MICCAI2024}. \\
\input{tables/primary_results}
% \input{tables/with_public}

% \subsection{Results:}
\noindent \textbf{Quantitative Results. }
Table~\ref{tab:perform} reports results across client configurations under both homogeneous and heterogeneous settings. In the homogeneous case, our method achieves the best performance in 6 configurations, with only a modest average gap of $0.37\%$ in the remaining 3. Under heterogeneous distributions, which represent a more realistic federated scenario, ProMoE-FL consistently outperforms all baselines across every configuration. Notably, PmcmFL shows the largest degradation, likely due to its reliance on a single class-wise prototype for missing modalities, which limits adaptability under distributional shifts.
Finally, in the heterogenous settings, we compare our method with CAR-MFL\cite{Pou_CARMFL_MICCAI2024}, SOTA for Public dataset based approach. The results demonstrate that our method achieves competitive performance.
We also ablate the mixture-of-experts component in the 8:0:2 heterogeneous setting. A single PCD achieves an AUC of 78.98, whereas the MoE variant attains 79.68, indicating a performance gain from incorporating expert routing.\\

\begin{figure}[t]
    \centering
    \includegraphics[width=0.95\linewidth]{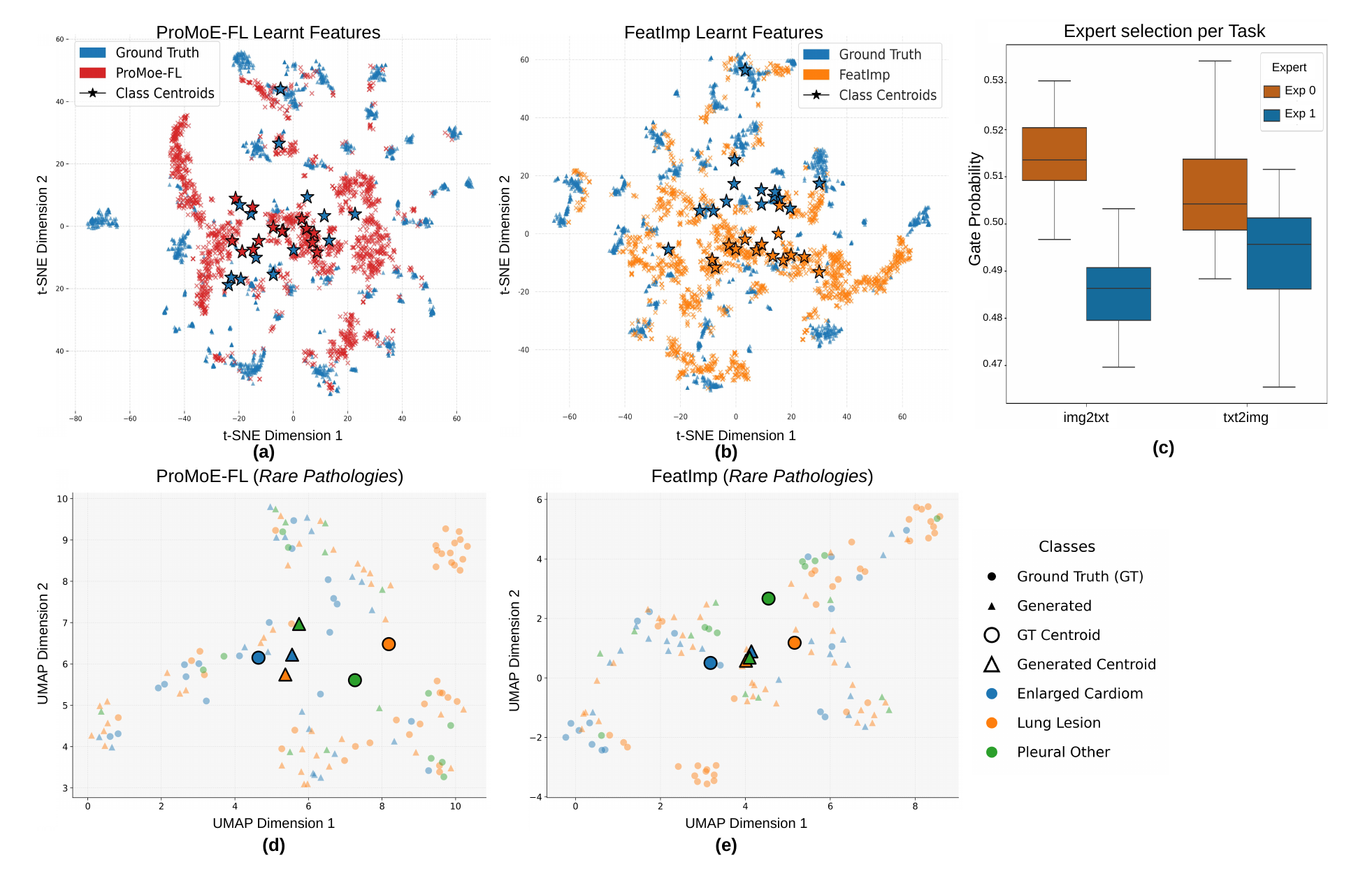}
    \caption{Qualitative Analysis of the methods. (a–b) t-SNE of learnt features with class centroids. (c) Gating distributions across experts. (d–e) UMAP of rare pathological classes.}
    \label{fig:qual}
\end{figure}

\noindent \textbf{Qualitative Results. }% \subsection{Qualitative Results}
Figure~\ref{fig:qual}(a–b) shows t-SNE analyses of synthesized features from the test set of \textcolor{Red}{ProMoE-FL}, \textcolor{Orange}{FeatImp} and the \textcolor{RoyalBlue}{ground truth}. 
% The qualitative t-SNE analysis 
FeatImp demonstrates mild centroid offsets and slightly more diffuse feature dispersion, indicating comparatively less precise alignment in the latent space.
In contrast, ProMoE-FL shows a tighter integration between synthesized and ground-truth features, with a more precise alignment of their class centroids. These observations suggest that ProMoE-FL provides a more refined and structurally faithful semantic mapping. 
As shown in Figure~\ref{fig:qual}(c), expert utilization remains balanced across both synthesis directions, with gating values centered near 0.5. This suggests stable multi-expert collaboration rather than collapse to a single expert, supporting the use of weighted gating instead of sparse Top-$k$ routing.\\
%MICCAI suggested 4th level heading
\noindent \textbf{Clinical Relevancy.}
%first to incorporate spanish; clinical relevancy
The UMAP visualizations in Figure~\ref{fig:qual}(d–e) highlight performance on rare pathological classes. While FeatImp captures the broader pathological structure, it exhibits a degree of mode collapse where synthesized centroids for different pathologies converge toward a single point. For conditions such as Enlarged Cardiomediastinum, Fracture, and Lung Lesion, ProMoE-FL maintains clear separation between class centroids while remaining closely aligned with the ground truth.
This ability to prevent feature overlap is vital in a medical context, as it ensures that synthesized data from missing-modality patients retains the discriminative signals necessary for reliable diagnosis.
Moreover, PadChest~\cite{BUSTOS2020101797Padchest} contains Spanish radiology reports, whereas the other datasets are in English. To the best of our knowledge, this is the first work to consider linguistic domain heterogeneity in federated learning with missing modalities, reflecting realistic cross-lingual clinical settings.

\section{Conclusion}
In this work, we introduced ProMoE-FL, a prototype-conditioned Mixture-of-Experts framework for multimodal federated learning under missing modality constraints. 
By conditioning the synthesis on a client-aware global prototype bank of missing modality, 
ProMoE-FL enables robust cross-modal feature synthesis addressing data heterogeneity in multimodal federated setups without relying on public data. 
Extensive experiments across various client configurations show consistent improvements over state-of-the-art methods, particularly in heterogeneous (non-IID) settings, where conventional imputation approaches struggle to generalize. % Qualitative and quantitative analyses further 
We demonstrate strong alignment between our synthesized and ground-truth feature distributions, preserving diagnostic structures, including rare pathologies such as lung lesions, enlarged cardiomediastinum, and fractures. By moving beyond simple feature imputation toward adaptive prototype-conditioned synthesis, ProMoE-FL provides a practical approach to addressing data heterogeneity in real-world clinical federated learning.

\begin{credits}
\subsubsection{\ackname} This work was supported as part of the “Swiss AI initiative” by a grant from the Swiss National Supercomputing Centre (CSCS) under project ID a168 on Alps.

% \subsubsection{\discintname}
% % It is now necessary to declare any competing interests or to specifically
% % state that the authors have no competing interests. Please place the
% % statement with a bold run-in heading in small font size beneath the
% % (optional) acknowledgments\footnote{If EquinOCS, our proceedings submission
% % system, is used, then the disclaimer can be provided directly in the system.},
% % for example: 
% The authors have no competing interests to declare that are
% relevant to the content of this article. 
% % Or: Author A has received research
% % grants from Company W. Author B has received a speaker honorarium from
% % Company X and owns stock in Company Y. Author C is a member of committee Z.
\end{credits}

%
% ---- Bibliography ----
%
% BibTeX users should specify bibliography style 'splncs04'.
% References will then be sorted and formatted in the correct style.
%
\bibliographystyle{splncs04}
\bibliography{references}

\end{document}

%% file: tables/primary_results.tex
\begin{table}[!t]
\centering
\setlength\tabcolsep{2pt}
\renewcommand{\arraystretch}{1.20}
\caption{AUC$\uparrow$ Performance in Homogeneous and Heterogeneous partitions. Client configurations are denoted by I:T:M, where I, T and M denote the number of Image-only, Text-Only and Multimodal Clients Respectively. Additionally, experiments conducted with public data (Pub.) are also included.}
\resizebox{1\columnwidth}{!}{
\begin{tabular}{|ll|c|c|c|c|c|c|c|c|c|}

\hline
   & \multirow{2}{*}{Method}  & \multicolumn{3}{c}{} & \multicolumn{3}{c}{Partitions (I:T:M)} & \multicolumn{3}{c|}{}\\ \cline{3-11}
   & & 4:0:6          & 0:4:6          & \multicolumn{1}{c|}{2:2:6}          & 6:0:4          & 0:6:4          & \multicolumn{1}{c|}{3:3:4}          & 8:0:2          & 0:8:2          & 4:4:2          \\ \hline
   
\multirow{5}{*}{\begin{sideways}Homogeneous\end{sideways}} & FedAvg~\cite{McMahan2017Communication} + Zero         & 86.64$_{\pm 0.46}$          & 87.95$_{\pm 0.53}$          & \multicolumn{1}{c|}{87.40$_{\pm 0.54}$}           & 82.99$_{\pm 1.03}$          & 87.01$_{\pm 0.33}$          & \multicolumn{1}{c|}{84.48$_{\pm 0.61}$}          & 79.69$_{\pm 1.61}$          & 86.32$_{\pm 0.52}$          & 80.47$_{\pm 0.56}$          \\
   & FedAvg~\cite{McMahan2017Communication} + Unif. &  87.79$_{\pm 0.63}$          & 89.27$_{\pm 0.56}$          & \multicolumn{1}{c|}{88.49$_{\pm 0.56}$}          & 84.83$_{\pm 0.32}$          &  88.64$_{\pm 0.25}$          & \multicolumn{1}{c|}{ 86.90$_{\pm 0.25}$}          &  80.60$_{\pm 0.91}$          &  88.37$_{\pm 0.34}$          & 85.74$_{\pm 0.28}$         \\
   & FeatImp~\cite{poudel2025multimodalfederatedlearningmissing}         & 89.31$_{\pm 0.33}$          & \textbf{89.52$_{\pm 0.11}$}          & \multicolumn{1}{c|}{89.12$_{\pm 0.55}$}          & 87.61$_{\pm 0.12}$          & \textbf{89.30$_{\pm 0.42}$}          & \multicolumn{1}{c|}{88.12$_{\pm 0.47}$}          & 86.16$_{\pm 0.44}$          & \textbf{88.98$_{\pm 0.28}$}          & 88.79$_{\pm 0.12}$          \\
   & PmcmFL~\cite{bao2024multimodalfederatedlearningmissing}         & 84.54$_{\pm 0.17}$ & 83.76$_{\pm 0.92}$ & \multicolumn{1}{c|}{84.78$_{\pm 0.24}$} & 81.72$_{\pm 0.38}$ & 84.21$_{\pm 0.46}$ & \multicolumn{1}{c|}{83.07$_{\pm 0.21}$} & 77.03$_{\pm 0.70}$ & 81.19$_{\pm 0.50}$ & 82.56$_{\pm 0.24}$ \\ 
   
   & ProMoE-FL (ours)         & \textbf{89.44$_{\pm 0.20}$} & 89.17$_{\pm 0.43}$ & \multicolumn{1}{c|}{\textbf{89.41$_{\pm 0.20}$}} & \textbf{88.27$_{\pm 0.75}$} & 89.13$_{\pm 0.29}$ & \multicolumn{1}{c|}{\textbf{89.25$_{\pm 0.27}$}} & \textbf{87.46$_{\pm 0.47}$} & 88.51$_{\pm 0.58}$ & \textbf{89.41$_{\pm 0.62}$} \\ \hline
   
\hline
   % & &  \multicolumn{3}{c}{} & \multicolumn{3}{c}{Heterogenous} & \multicolumn{3}{c|}{}\\ \hline
   % & (I:T:M)         & 4:0:6          & 0:4:6          & \multicolumn{1}{c|}{2:2:6}          & 6:0:4          & 0:6:4          & \multicolumn{1}{c|}{3:3:4}          & 8:0:2          & 0:8:2          & 4:4:2          \\ \hline
   
\multirow{5}{*}{\begin{sideways}Heterogeneous\end{sideways}} & FedAvg~\cite{McMahan2017Communication} + Zero & 79.32$_{\pm 1.24}$	&74.93$_{\pm 0.86}$	& \multicolumn{1}{c|}{76.30$_{\pm 1.06}$}	&78.95$_{\pm 0.55}$	&74.41$_{\pm 0.33}$	&\multicolumn{1}{c|}{74.98$_{\pm 0.70}$}	&72.76$_{\pm 0.78}$	&73.34$_{\pm 0.86}$	&73.25$_{\pm 0.30}$ \\

% & 73.34          & 74.407          & \multicolumn{1}{c|}{74.931}           & 73.246          & 74.985          & \multicolumn{1}{c|}{76.299}          & 72.76          & 78.951          & 79.323          \\

   & FedAvg~\cite{McMahan2017Communication} + Unif. &78.55$_{\pm 1.88}$&	76.58$_{\pm 0.63}$	& \multicolumn{1}{c|}{77.74$_{\pm 1.00}$} &	77.35$_{\pm 0.64}$ &	76.48$_{\pm 1.01}$ &	\multicolumn{1}{c|}{77.81$_{\pm 0.77}$} &	71.59$_{\pm 0.28}$ &	75.63$_{\pm 1.07}$ &	76.69$_{\pm 0.81}$ \\

   % & 75.63          & 76.476          & \multicolumn{1}{c|}{76.581}          & 76.695          & 77.812          & \multicolumn{1}{c|}{77.744}          & 71.59          & 77.353          & 78.546          \\
   
   & FeatImp~\cite{poudel2025multimodalfederatedlearningmissing}   & 80.19$_{\pm 0.92}$&	76.09$_{\pm 0.37}$&	\multicolumn{1}{c|}{78.42$_{\pm 0.62}$} &	80.93$_{\pm 0.53}$ &	77.14$_{\pm 0.59}$ &	\multicolumn{1}{c|}{78.56$_{\pm 1.49}$} &	77.15$_{\pm 1.34}$ & 	76.78$_{\pm 0.65}$ &	80.06$_{\pm 1.45}$ \\
   
   % & 76.78          & 77.5          & \multicolumn{1}{c|}{77.56}          & 80.06          & 79.4          & \multicolumn{1}{c|}{78.61}          & 77.15          & 80.93          & 80.187          \\
   
   & PmcmFL~\cite{bao2024multimodalfederatedlearningmissing}     & 74.66$_{\pm 0.99}$&	72.10$_{\pm 0.27}$&	\multicolumn{1}{c|}{76.99$_{\pm 0.45}$} &	71.01$_{\pm 0.40}$ &	68.39$_{\pm 0.16}$ &	\multicolumn{1}{c|}{75.96$_{\pm 1.27}$} &	67.51$_{\pm 0.79}$ &	64.65$_{\pm 0.51}$ &	71.48$_{\pm 0.67}$ \\

   & ProMoE-FL (ours)  & \textbf{81.91$_{\pm 0.83}$} &	\textbf{78.28$_{\pm 0.32}$} &	\multicolumn{1}{c|}{\textbf{79.88$_{\pm 0.43}$}} &	\textbf{83.02$_{\pm 0.26}$} &	\textbf{77.96$_{\pm 1.54}$} &	\multicolumn{1}{c|}{\textbf{80.29$_{\pm 1.35}$}}	& \textbf{79.68$_{\pm 1.52}$} &	\textbf{78.18$_{\pm 0.07}$} &	\textbf{80.96$_{\pm 0.72}$} \\
   
   % & \textbf{77.28} & \textbf{78.21} & \multicolumn{1}{c|}{\textbf{78.02}} & \textbf{80.96} & \textbf{81.37} & \multicolumn{1}{c|}{\textbf{79.92}} & \textbf{79.68} & \textbf{83.02} & \textbf{81.914} \\
   
   \hline
    \hline
   \multirow{2}{*}{\begin{sideways}Pub.\end{sideways}}&CAR-MFL~\cite{Pou_CARMFL_MICCAI2024} &  87.87$_{\pm 0.12}$          & 87.65$_{\pm 0.31}$          & \multicolumn{1}{c|}{\textbf{88.95$_{\pm 0.33}$}}          & 87.66$_{\pm 0.45}$          &  86.61$_{\pm 0.14}$          & \multicolumn{1}{c|}{ 88.29$_{\pm 0.55}$}          &  85.93$_{\pm 0.24}$          &  \textbf{86.32$_{\pm 0.54}$}          & 88.19$_{\pm 0.17}$         \\

       &ProMoE-FL (ours)         & \textbf{89.16$_{\pm 0.36}$} & \textbf{87.76$_{\pm 0.01}$} & \multicolumn{1}{c|}{87.98$_{\pm 1.50}$} & \textbf{89.79$_{\pm 0.09}$} & \textbf{88.10$_{\pm 0.71}$} & \multicolumn{1}{c|}{\textbf{88.86$_{\pm 0.57}$}} & \textbf{89.28$_{\pm 0.51}$} & 85.90$_{\pm 0.33}$ & \textbf{89.19$_{\pm 0.68}$} \\ \hline

\end{tabular}
}
\label{tab:perform}
\end{table}

%% file: references.bib
@article{Acosta2022Multimodal,
  author    = {Acosta, J.N. and Falcone, G.J. and Rajpurkar, P. and Topol, E.J.},
  title     = {Multimodal biomedical AI},
  journal   = {Nature Medicine},
  volume    = {28},
  number    = {9},
  pages     = {1773--1784},
  year      = {2022}
}

@article{Shrestha2023Medical,
  author    = {Shrestha, P. and Amgain, S. and Khanal, B. and Linte, C.A. and Bhattarai, B.},
  title     = {Medical vision language pretraining: A survey},
  journal   = {arXiv preprint arXiv:2312.06224},
  year      = {2023}
}

@article{jmirmultimodal,
author="AlSaad, Rawan
and Abd-alrazaq, Alaa
and Boughorbel, Sabri
and Ahmed, Arfan
and Renault, Max-Antoine
and Damseh, Rafat
and Sheikh, Javaid",
title="Multimodal Large Language Models in Health Care: Applications, Challenges, and Future Outlook",
journal="J Med Internet Res",
year="2024",
month="Sep",
day="25",
volume="26",
pages="e59505",
issn="1438-8871"
}

@inproceedings{McMahan2017Communication,
  author    = {McMahan, B. and Moore, E. and Ramage, D. and Hampson, S. and y Arcas, B.A.},
  title     = {Communication-efficient learning of deep networks from decentralized data},
  booktitle = {Proceedings of the 20th International Conference on Artificial Intelligence and Statistics (AISTATS)},
  pages     = {1273--1282},
  publisher = {PMLR},
  year      = {2017}
}

@article{Sachin2023AMC,
  title={A Multimodal Contrastive Federated Learning for Digital Healthcare},
  author={D. N. Sachin and B. Annappa and Sateesh Ambasange and Alan E. Tony},
  journal={SN Computer Science},
  year={2023},
  volume={4},
  pages={1-12}
}

@ARTICLE{9891834qayyum,
  author={Qayyum, Adnan and Ahmad, Kashif and Ahsan, Muhammad Ahtazaz and Al-Fuqaha, Ala and Qadir, Junaid},
  journal={IEEE Open Journal of the Computer Society}, 
  title={Collaborative Federated Learning for Healthcare: Multi-Modal COVID-19 Diagnosis at the Edge}, 
  year={2022},
  volume={3},
  number={},
  pages={172-184}}

@article{chenReportgen,
    author = {Chen, Jieying and Pan, Rong},
    title = {Medical report generation based on multimodal federated learning},
    journal = {Computerized medical imaging and graphics : the official journal of the Computerized Medical Imaging Society},
    year = {2024},
    volume = {113},
    pages = {102342}
}

@article{chhetri2026medmmflmultimodalfederatedlearning,
      title={Med-MMFL: A Multimodal Federated Learning Benchmark in Healthcare}, 
      author={Aavash Chhetri and Bibek Niroula and Pratik Shrestha and Yash Raj Shrestha and Lesley A Anderson and Prashnna K Gyawali and Loris Bazzani and Binod Bhattarai},
      year={2026},
      journal = {arXiv preprint arXiv:2602.04416}
}

@article{Thrasher2023Multimodal,
  author    = {Thrasher, J. and Devkota, A. and Siwakotai, P. and Chivukula, R. and Poudel, P. and Hu, C. and Bhattarai, B. and Gyawali, P.},
  title     = {Multimodal federated learning in healthcare: A review},
  journal   = {arXiv preprint arXiv:2310.09650},
  year      = {2023}
}

@inbook{Seibold_2022,
   title={Breaking with Fixed Set Pathology Recognition Through Report-Guided Contrastive Training},
   ISBN={9783031164439},
   ISSN={1611-3349},
   booktitle={Medical Image Computing and Computer Assisted Intervention – MICCAI 2022},
   publisher={Springer Nature Switzerland},
   author={Seibold, Constantin and Reiß, Simon and Sarfraz, M. Saquib and Stiefelhagen, Rainer and Kleesiek, Jens},
   year={2022},
   pages={690–700} }

@inbook{You_2023,
   title={CXR-CLIP: Toward Large Scale Chest X-ray Language-Image Pre-training},
   ISBN={9783031438950},
   ISSN={1611-3349},
   booktitle={Medical Image Computing and Computer Assisted Intervention – MICCAI 2023},
   publisher={Springer Nature Switzerland},
   author={You, Kihyun and Gu, Jawook and Ham, Jiyeon and Park, Beomhee and Kim, Jiho and Hong, Eun K. and Baek, Woonhyuk and Roh, Byungseok},
   year={2023},
   pages={101–111} }

@misc{chen2023generativetextguided3dvisionlanguage,
      title={Generative Text-Guided 3D Vision-Language Pretraining for Unified Medical Image Segmentation}, 
      author={Yinda Chen and Che Liu and Wei Huang and Sibo Cheng and Rossella Arcucci and Zhiwei Xiong},
      year={2023},
      eprint={2306.04811},
      archivePrefix={arXiv},
      primaryClass={cs.CV}
}

@inproceedings{lee2024visionlanguagegenerativemodelviewspecific,
      title={Vision-Language Generative Model for View-Specific Chest X-ray Generation}, 
      author={Hyungyung Lee and Da Young Lee and Wonjae Kim and Jin-Hwa Kim and Tackeun Kim and Jihang Kim and Leonard Sunwoo and Edward Choi},
      booktitle={CHIL},
      year={2024}
}

@article{lau2019unifiedrepresentation,
      title={A unified representation network for segmentation with missing modalities}, 
      author={Kenneth Lau and Jonas Adler and Jens Sjölund},
      year={2019},
      journal={arXiv preprint arXiv:1908.06683}
}

@inproceedings{MMIN,
    title = "Missing Modality Imagination Network for Emotion Recognition with Uncertain Missing Modalities",
    author = "Zhao, Jinming  and
      Li, Ruichen  and
      Jin, Qin",
    editor = "Zong, Chengqing  and
      Xia, Fei  and
      Li, Wenjie  and
      Navigli, Roberto",
    booktitle = "ACL",
    month = aug,
    year = "2021",
    pages = "2608--2618"
}

@inproceedings{ShaSpec,
      title={Multi-modal Learning with Missing Modality via Shared-Specific Feature Modelling}, 
      author={Hu Wang and Yuanhong Chen and Congbo Ma and Jodie Avery and Louise Hull and Gustavo Carneiro},
      year={2024},
      booktitle={CVPR},
}

@inproceedings{SMIL,
      title={SMIL: Multimodal Learning with Severely Missing Modality}, 
      author={Mengmeng Ma and Jian Ren and Long Zhao and Sergey Tulyakov and Cathy Wu and Xi Peng},
      year={2021},
      booktitle={AAAI}
}

@inproceedings{bao2024multimodalfederatedlearningmissing,
      title={Multimodal Federated Learning with Missing Modality via Prototype Mask and Contrast}, 
      author={Guangyin Bao and Qi Zhang and Duoqian Miao and Zixuan Gong and Liang Hu and Ke Liu and Yang Liu and Chongyang Shi},
      year={2024},
      booktitle={ICML}
}

@article{LE2025103219,
title = {Cross-modal prototype based multimodal federated learning under severely missing modality},
journal = {Information Fusion},
volume = {122},
pages = {103219},
year = {2025},
issn = {1566-2535},
author = {Huy Q. Le and Chu Myaet Thwal and Yu Qiao and Ye Lin Tun and Minh N.H. Nguyen and Eui-Nam Huh and Choong Seon Hong}
}

@inproceedings{yu2023multimodalfederatedlearningcontrastive,
      title={Multimodal Federated Learning via Contrastive Representation Ensemble}, 
      author={Qiying Yu and Yang Liu and Yimu Wang and Ke Xu and Jingjing Liu},
      year={2023},
      booktitle={ICLR}
}

@inproceedings{fedcolaSun,
author = {Sun, Guangyu and Mendieta, Matias and Dutta, Aritra and Li, Xin and Chen, Chen},
title = {Towards Multi-modal Transformers in Federated Learning},
year = {2024},
isbn = {978-3-031-72632-c3},
publisher = {Springer-Verlag},
address = {Berlin, Heidelberg},
booktitle = {ECCV},
pages = {229–246},
numpages = {18},
location = {Milan, Italy}
}

@InProceedings{Pou_CARMFL_MICCAI2024,
        author = { Poudel, Pranav and Shrestha, Prashant and Amgain, Sanskar and Shrestha, Yash Raj and Gyawali, Prashnna and Bhattarai, Binod},
        title = { { CAR-MFL: Cross-Modal Augmentation by Retrieval for Multimodal Federated Learning with Missing Modalities } },
        booktitle = {MICCAI},
        year = {2024}
}

@inproceedings{poudel2025multimodalfederatedlearningmissing,
      title={Multimodal Federated Learning With Missing Modalities through Feature Imputation Network}, 
      author={Pranav Poudel and Aavash Chhetri and Prashnna Gyawali and Georgios Leontidis and Binod Bhattarai},
      year={2025},
      booktitle={MIUA}
}

@article{saha2024examiningmodalityincongruitymultimodal,
      title={Examining Modality Incongruity in Multimodal Federated Learning for Medical Vision and Language-based Disease Detection}, 
      author={Pramit Saha and Divyanshu Mishra and Felix Wagner and Konstantinos Kamnitsas and J. Alison Noble},
      year={2024},
      journal={arXiv preprint arXiv:2402.05294}
}

@inproceedings{dai2025unbiased,
      title = {Unbiased Missing-modality Multimodal Learning},
      author = {Raiting Dai and Chenxi Li and Yandong Yan and Liso Mo and Ke Qin and Tao He},
      booktitle = {ICCV},
      year = {2025}
}

@article{Saha_Mishra_Wagner_Kamnitsas_Noble_2025, 
title={Incongruent Multimodal Federated Learning for Medical Vision and Language-based Multi-label Disease Detection}, 
volume={39}, 
number={27}, 
journal={AAAI}, 
author={Saha, Pramit and Mishra, Divyanshu and Wagner, Felix and Kamnitsas, Konstantinos and Noble, J. Alison}, year={2025}, month={Apr.}, pages={28331-28339} }

@inproceedings{he2016deep,
  title={Deep residual learning for image recognition},
  author={He, Kaiming and Zhang, Xiangyu and Ren, Shaoqing and Sun, Jian},
  booktitle={CVPR},
  pages={770--778},
  year={2016}
}

@article{devlin2018bert,
  title={Bert: Pre-training of deep bidirectional transformers for language understanding},
  author={Devlin, Jacob and Chang, Ming-Wei and Lee, Kenton and Toutanova, Kristina},
  journal={arXiv preprint arXiv:1810.04805},
  year={2018}
}

@article{kingma2014adam,
  title={Adam: A method for stochastic optimization},
  author={Kingma, Diederik P and Ba, Jimmy},
  journal={arXiv preprint arXiv:1412.6980},
  year={2014}
}

@article{johnson2019mimic,
  title={MIMIC-CXR, a de-identified publicly available database of chest radiographs with free-text reports},
  author={Johnson, Alistair EW and Pollard, Tom J and Berkowitz, Seth J and Greenbaum, Nathaniel R and Lungren, Matthew P and Deng, Chih-ying and Mark, Roger G and Horng, Steven},
  journal={Scientific data},
  volume={6},
  number={1},
  pages={317},
  year={2019},
  publisher={Nature Publishing Group UK London}
}

@article{demner2016preparing,
  title={Preparing a collection of radiology examinations for distribution and retrieval},
  author={Demner-Fushman, Dina and Kohli, Marc D and Rosenman, Marc B and Shooshan, Sonya E and Rodriguez, Laritza and Antani, Sameer and Thoma, George R and McDonald, Clement J},
  journal={Journal of the American Medical Informatics Association},
  volume={23},
  number={2},
  pages={304--310},
  year={2016},
  publisher={Oxford University Press}
}

@inproceedings{irvin2019chexpert,
  title={Chexpert: A large chest radiograph dataset with uncertainty labels and expert comparison},
  author={Irvin, Jeremy and Rajpurkar, Pranav and Ko, Michael and Yu, Yifan and Ciurea-Ilcus, Silviana and Chute, Chris and Marklund, Henrik and Haghgoo, Behzad and Ball, Robyn and Shpanskaya, Katie and others},
  booktitle={AAAI},
  year={2019}
}

@article{BUSTOS2020101797Padchest,
title = {PadChest: A large chest x-ray image dataset with multi-label annotated reports},
journal = {Medical Image Analysis},
volume = {66},
pages = {101797},
year = {2020},
issn = {1361-8415},
author = {Aurelia Bustos and Antonio Pertusa and Jose-Maria Salinas and Maria {de la Iglesia-Vayá}}
}

@article{goldberger2000physionet,
  author    = {Goldberger, A. and Amaral, L. and Glass, L. and Hausdorff, J. and Ivanov, P. C. and Mark, R. and Stanley, H. E.},
  title     = {PhysioBank, PhysioToolkit, and PhysioNet: Components of a new research resource for complex physiologic signals},
  journal   = {Circulation [Online]},
  volume    = {101},
  number    = {23},
  pages     = {e215--e220},
  year      = {2000},
}

@inproceedings{li2021modelcontrastivefederatedlearning,
      title={Model-Contrastive Federated Learning}, 
      author={Qinbin Li and Bingsheng He and Dawn Song},
      year={2021},
      booktitle={CVPR}
}
